\documentclass[journal]{IEEEtran}
\usepackage{amsmath,amsfonts}
\usepackage{algorithmic}
\usepackage{algorithm}
\usepackage{array}
\usepackage[caption=false,font=normalsize,labelfont=sf,textfont=sf]{subfig}
\usepackage{textcomp}
\usepackage{stfloats}
\usepackage{url}
\usepackage{verbatim}
\usepackage{graphicx}
\usepackage{cite}
\usepackage{booktabs} 
\usepackage{multirow}
\usepackage{multicol}
\usepackage{makecell}
\usepackage{xcolor}
\usepackage{hyperref}

\begin{document}

\title{Distance Guided Generative Adversarial Network for Explainable Binary Classifications}

\author{Xiangyu Xiong,~\IEEEmembership{Member,~IEEE,} Yue Sun, Xiaohong Liu, Wei Ke, Chan-Tong Lam, Jiangang Chen,\\ Mingfeng Jiang, Mingwei Wang, Hui Xie, Tong Tong, Qinquan Gao, Hao Chen, Tao Tan
\thanks{This work is supported in part by the Macao Polytechnic University Grant (RP/FCA-05/2022) and in part by the Science and Technology Development Fund of Macao (0105/2022/A). (\emph{Corresponding author: Tao Tan.})}
\thanks{Xiangyu Xiong, Yue Sun, Chan-Tong Lam, Wei Ke, and Tao Tan are same with the Faculty of Applied Sciences, Macao Polytechnic University, Macao 999078, China (email: xiangyu.xiong.adc@gmail.com; yuesun@mpu.edu.mo; ctlam@mpu.edu.mo; wke@mpu.edu.mo; taotan@mpu.edu.mo).}
\thanks{Xiaohong Liu is with the John Hopcroft Center (JHC) for Computer Science, Shanghai Jiao Tong University, Shanghai 200240, China (email: xiaohongliu@sjtu.edu.cn).}
\thanks{Jiangang Chen is with the Shanghai Key Laboratory of Multidimensional Information Processing, School of Communication \& Electronic Engineering, East China Normal University, Shanghai 200241, China, and also with the Engineering Research Center of Traditional Chinese Medicine Intelligent Rehabilitation, Ministry of Education, Shanghai 201203, China (email: jgchen@cee.ecnu.edu.cn).}
\thanks{Mingfeng Jiang is with the School of Computer Sciecnes and Techonology, Zhejiang Sci-Tech University, Hangzhou 310018, Zhejiang, China (email: m.jiang@zstu.edu.cn). }
\thanks{Mingwei Wang is with the Department of Cardiology, Affiliated Hospital of Hangzhou Normal University, Hangzhou Institute of Cardiovascular Diseases, Hangzhou Normal University, Hangzhou 310015, China (email: wmw990556@hznu.edu.cn).}
\thanks{Hui Xie is with the Department of Radiation Oncology, Affiliated Hospital (Clinical College) of Xiangnan University, 423000, Chenzhou, Hunan, China (email: h.xie@xnu.edu.cn).}
\thanks{Tong Tong and Qinquan Gao are same with the College of Physics and Information Engineering, Fuzhou University, Fuzhou 350108, China (email: ttraveltong@gmail.com; gqinquan@fzu.edu.cn.com).}
\thanks{Hao Chen is with the Department of Mathware, Jiangsu JITRI Sioux Technologies Company, Ltd, Suzhou 215000, China (e-mail: hao.chen.gd@gmail.com).} }

\markboth{Journal of \LaTeX\ Class Files,~Vol.~14, No.~8, August~2021}%
{Shell \MakeLowercase{\textit{et al.}}: A Sample Article Using IEEEtran.cls for IEEE Journals}

\IEEEpubid{0000--0000/00\$00.00~\copyright~2021 IEEE}

\maketitle
\begin{abstract}
Despite the potential benefits of data augmentation for mitigating the data insufficiency, traditional augmentation methods primarily rely on the prior intra-domain knowledge. On the other hand, advanced generative adversarial networks (GANs) generate inter-domain samples with limited variety. These previous methods make limited contributions to describing the decision boundaries for binary classification. In this paper, we propose a distance guided GAN (DisGAN) which controls the variation degrees of generated samples in the hyperplane space. Specifically, we instantiate the idea of DisGAN by combining two ways. The first way is vertical distance GAN (VerDisGAN) where the inter-domain generation is conditioned on the vertical distances. The second way is horizontal distance GAN (HorDisGAN) where the intra-domain generation is conditioned on the horizontal distances. Furthermore, VerDisGAN can produce the class-specific regions by mapping the source images to the hyperplane. Experimental results show that DisGAN consistently outperforms the GAN-based augmentation methods with explainable binary classification. The proposed method can apply to different classification architectures and has potential to extend to multi-class classification. We provide the code in \href{https://github.com/yXiangXiong/DisGAN}{https://github.com/yXiangXiong/DisGAN}.
\end{abstract}


\section{Introduction}
Deep neural networks achieve excellent performance in the computer vision fields~\cite{DBLP:journals/jbd/AlzubaidiZHADAS21a}, such as image classification~\cite{DBLP:conf/nips/KrizhevskySH12}, object detection~\cite{yolo}, image segmentation~\cite{unet} and image registration~\cite{voxelmorph}. In all these fields, a large-scale dataset containing sufficient supervisory information is crucial for effectively training of neural networks. However, in many realistic scenarios, neural networks can only be trained on the small-scale datasets, resulting in overfitting on the training set and poor generalization on the testing set. Although regularization techniques, such as parameter norm penalties, dropout~\cite{dropout}, batch normalization~\cite{batchnormlization}, layer normalization~\cite{layernormalization} and group normalization~\cite{groupnormalization}, have been developed to prevent overfitting, data augmentation is another way to address the training samples insufficiency.
\begin{figure*}[tbp]
	\centering
	\includegraphics[width=0.9\textwidth]{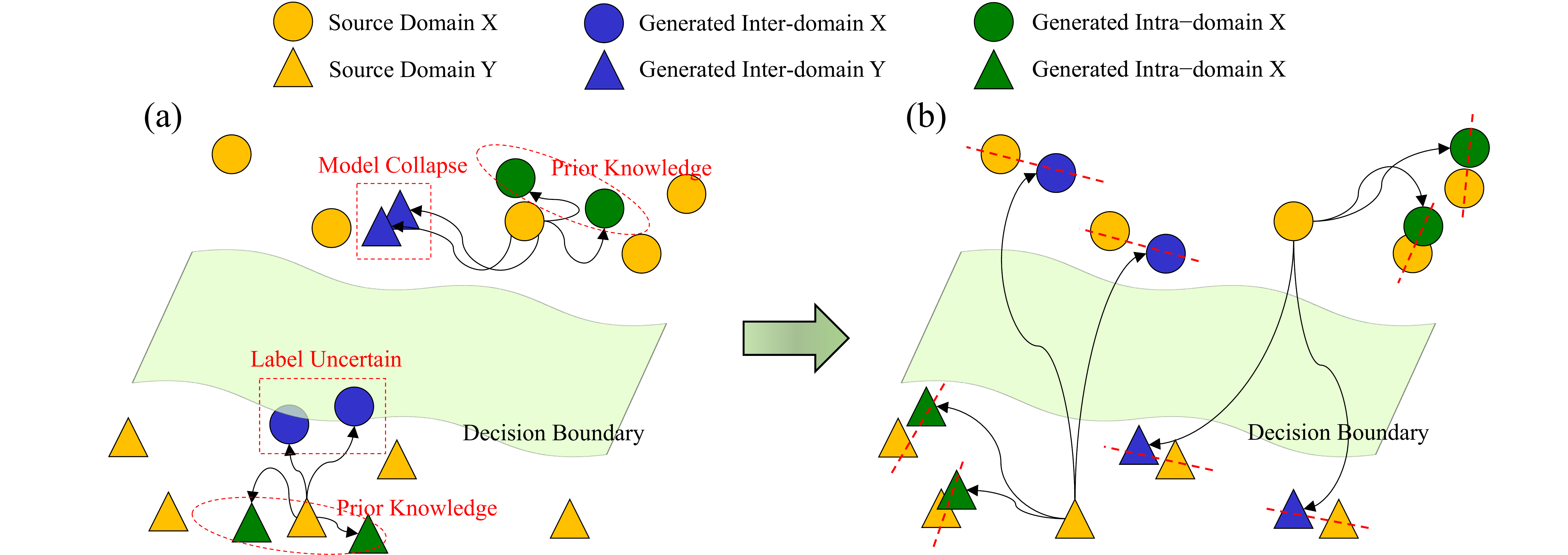}
	\caption{The illustration of data augmentation for binary classification under two settings: (a) Not controlling the variation degrees. The generated samples are intra-domain by traditional augmentation; and degenerate to be intra-domain due to model collapse and have uncertain domain labels by GAN-based augmentation. (b) Controlling the variation degrees. The generated samples can reshape the decision boundary in the hyperplane space.}
	\label{fig_Motivation}
\end{figure*}

Traditional augmentation method transforms the training samples by exploring prior intra-domain knowledge \cite{DBLP:conf/nips/KrizhevskySH12, High-Performance}, including random cropping, rotations, etc. However, these methods should be designed based on the specific scene. Other traditional augmentation techniques \cite{aug_feature, mixup} are proposed to weaken this limit. Generative adversarial network (GAN)~\cite{GAN} aims to generate the inter-domain samples which have the same distribution with the target domain's samples. However, due to the issue of mode collapsing in GANs \cite{ModeCollapse}, the generated samples' quality become uncertain and their variety is inferior to that of the real samples. Additionally, there is uncertainty regarding the domain labels of inter-domain generated samples. Here, domain implies a set of images that can be grouped as visually same category.
\IEEEpubidadjcol

Previous augmentation methods blindly augment the training samples. As illustrated in Figure~\ref{fig_Motivation}, we propose that controlling the variation degrees of generated samples can produce a more accurate decision boundary for the binary classifications. To achieve this, we develop a hyperplane distance GAN (DisGAN) which consists of vertical distance GAN (VerDisGAN) and horizontal distance GAN (HorDisGAN). Firstly, we train a binary classifier to construct an optimal hyperplane by using hinge loss~\cite{hinge_loss}. Secondly, we measure the two types distances: the vertical distances from the target inter-domain samples to the constructed hyperplane; and the horizontal distances from the target intra-domain samples to the source samples. Thirdly, we generate the inter-domain and intra-domain samples which are conditioned on the vertical distances and horizontal distances respectively. Experiments show that DisGAN consistently outperforms the GAN-based augmentation methods on the limited datasets. In addition, VerDisGAN can provide the class-difference map (CDM) for the binary classification. The CDMs enables a more transparent explanation than using Grad-CAM~\cite{gradcam}.

Our main contributions are listed as follows: (1) Training a binary classifier via hinge loss and fixing its weights to produce an optimal hyperplane. (2) A DisGAN enables controlling the variation degrees of generated samples. (3) An effective data augmentation algorithm which can reshape the decision boundaries for various classification architectures. (4) A class-difference map which provides the interpretability for binary classifications.

\section{Related Work}

\subsection{Image-to-Image Translation}
Recent works have achieved success in two-domain translation. For instance, Pix2pix~\cite{pix2pix} learns the general image translation in a supervised setting via L1+cGAN loss. However, it requires aligned image pairs due to the pixel-level reconstruction constraints. To alleviate the requirement of paired image supervision, unpaired two-domain translation networks have been proposed. UNIT~\cite{UNIT} is a coupled VAE-GAN algorithm based on a shared-latent space assumption. CycleGAN~\cite{cyclegan} and DiscoGAN~\cite{DiscoGAN} enforce bidirectional translation by utilizing a cycle consistency loss. In this study, the proposed DisGAN enables both the unpaired inter-domain and the intra-domain translation.

\subsection{GAN-based Augmentation Methods}
Recently, GAN~\cite{GAN} and its variations have been employed as the augmentation tools for the image classification. For instance, DCGAN~\cite{DCGAN} generates high-quality CT images for each liver lesion class~\cite{liver}. In this augmentation scheme, the generated inner-domain samples cannot well describe the decision boundary. Auxiliary classifiers have been combined with GANs to create soft labels for inter-domain generated data~\cite{HongjiangShi, ECGAN}. They alleviate the generated samples' authenticity by setting weights for them. However, soft labels are not precise to benefit the classification tasks, and finding appropriate classification weights for unreliable data is difficult. Other schemes generate inter-domain data via CycleGAN~\cite{Melanoma, LeafGAN, Ghazal}. Unfortunately, the generated samples close to the distribution of the source domain are given the target-domain labels. Unlike the above methods, the proposed DisGAN enforces the generated samples being close to the target samples, which contributes to reshaping the hyperplane.

\subsection{Image Generation Combining Classifiers}
Starting with cGAN~\cite{cGAN}, auxiliary information such as class labels has been integrated with GANs to generate samples of a given specific type. SGAN~\cite{SGAN} combines an auxiliary classifier with a discriminator to improve the generation performance. Several studies introduce an auxiliary classifier to reconstruct the auxiliary information from the generated samples, such as ACGAN~\cite{ACGAN} and VACGAN~\cite{VACGAN}. Later, this idea is expanded to inter-domain image translation by reconstructing target domain labels, such as conditional CycleGANs~\cite{AGcCyeleGAN, cCycleGAN} and StarGAN~\cite{stargan}. The class labels reconstructed in these methods cannot reflect the variation degrees of generated samples. Recently, ParaGAN~\cite{paraGAN} combines the pixel-level constraint and the distance-level constraint to control the changes of inter-domain generated samples. We further extend the distance constraint of ParaGAN.
\begin{figure*}[tbp]
	\centering
	\includegraphics[width=\textwidth]{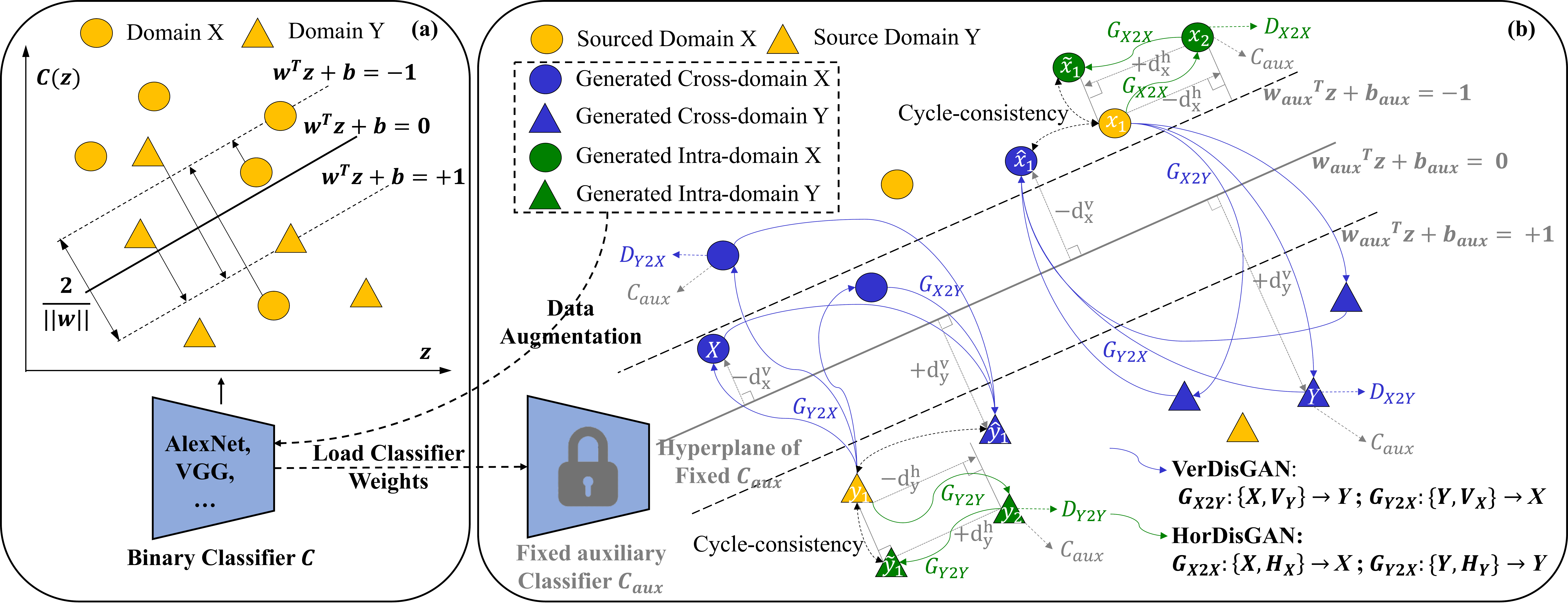}
	\caption{Overview of the proposed data augmentation method. (a) Training a binary classifier to obtain an optimal hyperplane via hinge loss. The fixed classifier is taken as an auxiliary classifier for image generation. (b) The DisGAN consists of VerDisGAN and HorDisGAN. They map the source images to the target images by taking the distances as input parameters. The auxiliary classifier is used to reconstruct the input distances from the generated images. In addition, they map the generated images back to source images. The reconstructed source images should be consist with the source images.}
	\label{fig_Architecture}
\end{figure*}

\section{Proposed Method}
\subsection{Problem Definition}
We control the variation degrees of the generated samples by integrating the distance space into the image space. Given the training samples $\left\{ x_i \right\}_{i=1}^N$ where ${x_i}\in{X}$, and $\left\{ y_j \right\}_{j=1}^M$ where ${y_j}\in{Y}$, we denote the data distribution $x\sim P_X(x)$ and $y \sim P_Y(y)$; hyperplane vertical distance $d_x^v\in V_X$ and $d_y^v\in V_Y$ and hyperplane horizontal distance $d_x^h\in H_X$ and $d_y^h\in H_Y$. We define four mappings $G_{X2Y}: \left\{X, V_Y\right\} \rightarrow Y$, $G_{Y2X}: \left\{Y, V_X\right\} \rightarrow X$, $G_{X2X}: \left\{X, H_X\right\} \rightarrow X$ and $G_{Y2Y}: \left\{Y, H_Y\right\} \rightarrow Y$, where the inter-domain mapping and the intra-domain mapping are conditioned on the vertical distances and the horizontal distances respectively. In addition, we introduce four adversarial discriminators $D_{X2Y}$, $D_{Y2X}$, $D_{X2X}$ and $D_{Y2Y}$, where $D_{X2Y}$ aims to discriminate between real images $\left\{ y \right\}$ and generated images $\left\{ G_{X2Y}(x, d_y^v) \right\}$; in the same way, $D_{Y2X}$ aims to discriminate between $\left\{ x \right\}$ and $\left\{ G_{Y2X}(y, d_x^v) \right\}$. $D_{X2X}$ aims to discriminate between real images $\left\{ x \right\}$ and $\left\{ G_{X2X}(x, d_x^h) \right\}$; in the same way,  $D_{Y2Y}$ aims to discriminate between $\left\{ y \right\}$ and $\left\{ G_{Y2Y}(y, d_y^h) \right\}$.

\subsection{Vertical and Horizontal Distances}
\subsubsection{Vertical Distance}
As shown in Fig \ref{fig_Architecture}(a), we train a binary classifier via hinge loss~\cite{hinge_loss} to obtain an optimal hyperplane which divides the training samples into two classes. Given a linear binary classifier $C(z) = w^Tz + b$ which is trained on a training set $\left\{ z_i, c_i \right\}_{i=1}^N$, $z_i \in R^D$, $c_i \in \left\{ -1, +1 \right\}$, the hinge loss can be expressed as follows:
\begin{equation}
\small
\mathcal{L}_{\rm hinge}(c_i, C(z_i)) =  \frac{1}{N}\sum_{n=1}^N max[0, 1-c_i(w^Tz_i + b))].
\label{hinge}
\end{equation}

The vertical distances are considering as the controllable parameter to enable inter-domain generation between domain $X$ and domain $Y$. To measure the vertical distances from target samples to the optimal hyperplane, we introduce an auxiliary classifier $C_{aux}(z) = w_{aux}^Tz+b_{aux}$ by fixing the binary classifier $C(z)$. Specifically, given a random sample $x \in X$ and a random target sample $y \in Y$, the vertical distances $d_v(x)$ and $d_v(y)$ from the samples to the optimal hyperplane ($w_{aux}^Tz+b_{aux}=0$) are defined as follows, respectively:
\begin{equation}
\small
d_v(x) = \lvert C_{aux}(x) \rvert = \lvert w_{aux}^Tx + b_{aux}\rvert,
\label{dvx}
\end{equation}
\begin{equation}
\small
d_v(y) = \lvert C_{aux}(y) \rvert = \lvert w_{aux}^Ty + b_{aux}\rvert.
\label{dvy}
\end{equation}

\subsubsection{Horizontal Distance}
The horizontal distances are considering as the controllable parameters to enable intra-domain generation in domain $X$ and domain $Y$ respectively. To measure the horizontal distances from the source samples to target samples, we firstly obtain the coordinate distances from the source samples to target samples. The coordinates are represented by the vectors extracted before the last fully connection layer of the auxiliary classifier $C_{aux}$. Specifically, given a random source sample $x_1 \in X$ and a random target sample $x_2 \in X$ with extracted vector $[{coor}_1^{x_1}, {coor}_2^{x_1}, \cdots, {coor}_m^{x_1}]$ and vector $[{coor}_1^{x_2}, {coor}_2^{x_2}, \cdots, {coor}_m^{x_2}]$ respectively, the coordinate distance $d_{coor}(x_1, x_2)$ between them is as follows:
\begin{equation}
\small
\begin{split}
d_{coor}(x_1, x_2) &= d_{vector}(C_{aux}(x_1), C_{aux}(x_2)) \\&= \sqrt{\sum_{i=1}^{m}({coor}_i^{x_1} - {coor}_i^{x_2})^2}, 
\end{split}
\end{equation}
Similarly, the coordinate distance between random source $y_1 \in Y$ and random target $y_2 \in Y$ is as follows:
\begin{equation}
\small
\begin{split}
d_{coor}(y_1, y_2) &= d_{vector}(C_{aux}(y_1), C_{aux}(y_2)) \\&= \sqrt{\sum_{i=1}^{m}({coor}_i^{y_1} - {coor}_i^{y_2})^2},
\end{split}
\end{equation}
We secondly obtain the vertical distances differences for the source samples and the target samples. Finally, we calculate the horizontal distances from the source samples to the target samples by using Pythagorean Theorem. Given the random samples $x_1, x_2 \in X$ and the random samples $y_1, y_2 \in Y$, the horizontal distances $d_h(x_1, x_2)$ and $d_h(y_1, y_2)$ are as follows, respectively:
\begin{equation}
\small
d_h(x_1, x_2)= \sqrt{d_{coor}(x_1, x_2)^2 - (d_v(x_1) - d_v(x_2))^2},
\label{dhx}
\end{equation}
\begin{equation}
\small
d_h(y_1, y_2) = \sqrt{d_{coor}(y_1, y_2)^2 - (d_v(y_1) - d_v(y_2))^2}.
\label{dhy}
\end{equation}

\subsection{VerDisGAN and HorDisGAN}
Figure~\ref{fig_Architecture}(b) illustrates the architecture of DisGAN.
\subsubsection{VerDisGAN}
Firstly, $G_{X2Y}$ translates a source sample $x_1$ into $G_{X2Y}(x_1, +d_v(y))$, and $D_{X2Y}$ distinguishes between this translated sample and a real sample $y$, and vice versa for $y_1$, $G_{Y2X}$ and $D_{Y2X}$. Secondly, an auxiliary classifier $C_{aux}$ reconstructs the vertical distances $+d_v(y)$ and $-d_v(x)$ from the inter-domain generated samples $G_{X2Y}(x_1, +d_v(y))$ and $G_{Y2X}(y_1, -d_v(x))$ respectively. Thirdly, $G_{Y2X}$ inversely translates $G_{X2Y}(x_1, +d_v(y))$ into the reconstructed sample $\hat{x}_1$ using the vertical distance $-d_v(x_1)$, and vice versa for obtaining the reconstructed $\hat{y}_1$.

\subsubsection{HorDisGAN} 
Firstly, $G_{X2X}$ transforms the source sample $x_1$ into $G_{X2X}(x_1, -d_h(x_1, x_2))$, and $D_{X2X}$ distinguishes between this translated sample and a real sample $x_2$, and vice versa for $y_1$, $G_{Y2Y}$ and $D_{Y2Y}$. Secondly, an auxiliary classifier $C_{aux}$ reconstructs the horizontal distances $-d_h(x_1, x_2)$ and $-d_h(y_1, y_2)$ from the intra-domain generated samples $G_{X2X}(x_1, -d_h(x_1, x_2))$ and $G_{Y2Y}(y_1, -d_h(y_1, y_2))$ respectively. Thirdly, $G_{X2X}$ inversely translates the $G_{X2X}(x_1, -d_h(x_1, x_2))$ into the reconstructed sample $\Tilde{x}_1$ using the horizontal distance $+d_h(x_1, x_2)$, and vice versa for obtaining the reconstructed $\Tilde{y}_1$.

\subsection{Objective Functions and Data Augmentation Algorithm}
We define the objective functions of VerDisGAN and HorDisGAN, respectively.

To make the generated samples be indistinguishable from the target samples, we apply LSGAN loss~\cite{lsgan} to inter-domain and intra-domain mappings. The LSGAN losses for $G_{X2Y}$ and $G_{X2X}$ are expressed as follows, respectively:
\begin{equation}
\small
\begin{split}
\mathcal{L}_{\rm interGAN}(&G_{X2Y}, D_{X2Y}, X, Y) \\&= \mathbb{E}_{y\sim P_Y}[(D_{X2Y}(y) - 1)^2] \\& + \mathbb{E}_{x_1\sim P_X}[D_{X2Y}(G_{X2Y}(x_1, +d_v(y)))^2],
\end{split}
\end{equation}
\begin{equation}
\small
\begin{split}
\mathcal{L}_{\rm intraGAN}(&G_{X2X}, D_{X2X}, X) \\&= \mathbb{E}_{x_2\sim P_X}[(D_{X2X}(x_2) - 1)^2] \\& + \mathbb{E}_{x_1\sim P_X}[D_{X2X}(G_{X2X}(x_1, -d_h(x_1, x_2)))^2].
\end{split}
\end{equation}
and vice versa for the inter-domain mapping $G_{Y2X}$ and the intra-domain mapping $G_{Y2Y}$.

To control the variation degrees of generated samples to be close as that of target samples, we design the distance losses for generated inter-domain and intra-domain samples according to Equations \ref{dvx}, \ref{dvy} and \ref{dhx}, \ref{dhy}. The two types distance losses are formulated as follows respectively:
\begin{equation}
\footnotesize
\begin{split}
\mathcal{L}_{\rm verDIS}(&G_{X2Y}, G_{Y2X}, C_{aux}) \\&= \mathbb{E}_{x_1\sim P_X}[\lVert \lvert d_v(y)\rvert-\lvert d_v(G_{X2Y}(x_1, +d_v(y)))\rvert \rVert _2^2] \\& + \mathbb{E}_{y_1\sim P_Y}[\lVert \lvert d_v(x)\rvert - \lvert d_v(G_{Y2X}(y_1, -d_v(x)))\rvert \rVert_2^2],
\end{split}
\end{equation}
\begin{equation}
\footnotesize
\begin{split}
&\mathcal{L}_{\rm horDIS}(G_{X2X}, G_{Y2Y}, C_{aux}) \\& = \mathbb{E}_{x_1\sim P_X}[\lVert d_h(x_1, x_2) - d_h(x_1, G_{X2X}(x_1, -d_h(x_1, x_2)))\rVert_2^2] \\& + \mathbb{E}_{y_1\sim P_Y}[\lVert d_h(y_1, y_2) - d_h(y_1, G_{Y2Y}(y_1, -d_h(y_1, y_2)))\rVert_2^2].
\end{split}
\end{equation}
\begin{algorithm}[tbp]
	\caption{Data augmentation with DisGAN samples}
	\label{alg1}
	\begin{algorithmic} 
		\REQUIRE Training samples $\left\{ x_i, -1\right\}_{i=1}^N$, $\left\{ y_j, +1 \right\}_{j=1}^M$, where $x_i \in X$ and $y_j \in Y$. Initialization: $C(z)$, $C'(z)$, $G_{X2Y}$, $D_{X2Y}$, $G_{Y2X}$, $D_{Y2X}$, $G_{X2X}$, $D_{X2X}$, $G_{Y2Y}$ and $D_{Y2Y}$.
        \STATE $\bullet$ Step 1:
        \STATE -Train binary classifier $C(z) = w^Tz + b$ by using hinge loss $\mathcal{L}_{\rm hinge}(-1, C(x_i))$ and $\mathcal{L}_{\rm hinge}(+1, C(y_j))$.
        \STATE $\bullet$ Step 2:
        \STATE -Fix $C(z)$ to get an auxiliary classifier $C_{aux}(z) = w_{aux}^Tz+b_{aux}$ and the optimal hyperplane $w_{aux}^Tz+b_{aux}=0$.
        \STATE $\bullet$ Step 3:
        \FOR{ $i=1$ to $\max \left\{M, N \right\}$ }
        \STATE -Let $i \leftarrow i\mod M$. Get random $l \leftarrow l\mod M$, $j \leftarrow j\mod N$ and $k \leftarrow k\mod N$. $ i,j,k \leq \max \left\{M, N \right\}$.
        \STATE -Sample a sample $x_i \in X$, a random sample $x_l
        \in X$ and random target samples $y_j, y_k \in Y$.
        \STATE -Measure vertical distances $d_v(x_i)$, $d_v(y_j)$, and horizontal distances $d_h(x_i, x_l)$, $d_h(y_j, y_k)$.
        \STATE -Generate the inter-domain samples: $G_{X2Y}(x_i, +d_v(y_j))$, $G_{Y2X}(y_j, -d_v(x_i))$ and the intra-domain samples: $G_{X2X}(x_i, -d_h(x_i, x_l))$, $G_{Y2Y}(y_j, -d_h(y_j, y_k))$.
        \STATE -Reconstruct the vertical and horizontal distances from the generated inter-domain and intra-domain samples using $C_{aux}$ respectively.
        \STATE -Inversely translate the generated samples to reconstruct the source images.
        \STATE -Update the generators using LSGAN loss, distance loss and cycle-consistency loss.
        \STATE -Determine the generated samples are real or fake using the discriminators $D_{X2Y}$, $D_{Y2X}$, $D_{X2X}$ and $D_{Y2Y}$.
        \STATE -Update the discriminators using LSGAN loss.
        \STATE -Update the classifier $C'(z)$ with the source samples and all generated samples by using hinge loss.
        \ENDFOR
	\end{algorithmic} 
\end{algorithm}

To make samples not lose the original information after translating twice, we develop the cycle-consistency losses \cite{cyclegan, stargan} for the inter-domain and intra-domain generation:
\begin{equation}
\scriptsize
\begin{split}
&\mathcal{L}_{\rm interCYC}(G_{X2Y}, G_{Y2X}) \\&= \mathbb{E}_{x_1\sim P_X}[||x_1 - G_{Y2X}(G_{X2Y}(x_1, +d_v(y)), -d_v(x_1))||_1] \\& + \mathbb{E}_{y_1\sim P_Y}[||y_1 -G_{X2Y}(G_{Y2X}(y_1, -d_v(x)), +d_v(y_1))||_1],
\end{split}
\end{equation}
\begin{equation}
\scriptsize
\begin{split}
&\mathcal{L}_{\rm intraCYC}(G_{X2X}, G_{Y2Y}) =\\& \mathbb{E}_{x_1\sim P_X}[||x_1 - G_{X2X}(G_{X2X}(x_1, -d_h(x_1, x_2)), +d_h(x_1, x_2))||_1] \\& + \mathbb{E}_{y_1\sim P_Y}[||y_1 -G_{Y2Y}(G_{Y2Y}(y_1, -d_h(y_1, y_2)), +d_h(y_1, y_2))||_1].
\end{split}
\end{equation}

Finally, the objective functions of DisGAN for inter-domain generation and intra-domain generation are formulated as follows, respectively:
\begin{equation}
\small
\begin{split}
\mathcal{L}_{\rm VerDisGAN}(&G_{X2Y}, G_{Y2X}, C_{aux}, D_{X2Y}, D_{Y2X}) \\& = \mathcal{L}_{\rm interGAN}(G_{X2Y}, D_{X2Y}, X, Y) \\&+ \mathcal{L}_{\rm interGAN}(G_{Y2X}, D_{Y2X}, Y, X) \\& + \lambda_{\rm verDIS}\mathcal{L}_{\rm verDIS}(G_{X2Y}, G_{Y2X}, C_{aux}) \\& + \lambda_{\rm crossCYC}\mathcal{L}_{\rm crossCYC}(G_{X2Y}, G_{Y2X}),
\end{split}
\end{equation}
\begin{equation}
\small
\begin{split}
\mathcal{L}_{\rm HorDisGAN}(&G_{X2X}, G_{Y2Y}, C_{aux}, D_{X2X}, D_{Y2Y}) \\& = \mathcal{L}_{\rm intraGAN}(G_{X2X}, D_{X2X}, X, Y) \\&+ \mathcal{L}_{\rm intraGAN}(G_{Y2Y}, D_{Y2Y}, Y, X) \\& + \lambda_{\rm horDIS}\mathcal{L}_{\rm horDIS}(G_{X2Y}, G_{Y2X}, C_{aux}) \\& + \lambda_{\rm intraCYC}\mathcal{L}_{\rm intraCYC}(G_{X2X}, G_{Y2Y}).
\end{split}
\end{equation}
where $\lambda_{\rm verDIS}, \lambda_{\rm interCYC}, \lambda_{\rm horDIS}, \lambda_{\rm intraCYC} > 0$ are some hyper-parameters balancing the losses.

The data augmentation is shown in the Algorithm \ref{alg1}.
\begin{table*}[tbp]
  \centering
  \scriptsize
  \caption{Hyper-parameters settings of the $\lambda_{\rm verDIS}$ and the $\lambda_{\rm horDIS}$ for DisGAN.}
    \begin{tabular}{ccccccccccc}
    \toprule
    \multicolumn{2}{c}{Datasets} & Hyper-parameters & \multicolumn{1}{c}{AlexNet} & \multicolumn{1}{c}{VGG16} & \multicolumn{1}{c}{GoogLeNet} & \multicolumn{1}{c}{ResNet34} & \multicolumn{1}{c}{DenseNet121} & \multicolumn{1}{c}{MnasNet1\_0} & \multicolumn{1}{c}{EfficientNet  V1} & \multicolumn{1}{c}{ConvNeXt} \\
    \midrule
    \multicolumn{2}{c}{\multirow{2}[4]{*}{Butterfly Mimics}} & $\lambda_{\rm verDIS}$ & 0.1   & 0.1   & 0.1   & 0.1   & 0.1   & 0.1   & 0.1   & 0.1 \\
\cmidrule{3-11}    \multicolumn{2}{c}{} & $\lambda_{\rm horDIS}$ & 0.001 & 0.001 & 0.01  & 0.001 & 0.001 & 0.001 & 0.01  & 0.01 \\
    \midrule
    \multicolumn{2}{c}{\multirow{2}[4]{*}{\makecell{Asian vs African \\ Elephants}}} & $\lambda_{\rm verDIS}$ & 0.1   & 0.1   & 0.1   & 0.1   & 0.1   & 0.1   & 0.1   & 0.1 \\
\cmidrule{3-11}    \multicolumn{2}{c}{} & $\lambda_{\rm horDIS}$ & 0.001 & 0.001 & 0.01  & 0.01  & 0.001 & 0.001 & 0.01  & 0.01 \\
    \midrule
    \multicolumn{2}{c}{\multirow{2}[4]{*}{Breast Ultrasound}} & $\lambda_{\rm verDIS}$ & 0.1   & 0.01  & 0.1   & 0.1   & 0.01  & 0.1   & 0.01  & 0.1 \\
\cmidrule{3-11}    \multicolumn{2}{c}{} & $\lambda_{\rm horDIS}$ & 0.001 & 0.001 & 0.001 & 0.001 & 0.001 & 0.001 & 0.001 & 0.01 \\
    \midrule
    \multicolumn{2}{c}{\multirow{2}[4]{*}{COVID-CT}} & $\lambda_{\rm verDIS}$ & 0.1   & 0.01  & 0.1   & 0.1   & 0.01  & 0.1   & 0.1   & 0.1 \\
\cmidrule{3-11}    \multicolumn{2}{c}{} & $\lambda_{\rm horDIS}$ & 0.001 & 0.001 & 0.001 & 0.001 & 0.001 & 0.001 & 0.01  & 0.01 \\
    \bottomrule
    \end{tabular}%
\label{tab_hyperparameters}
\end{table*}%

\begin{table*}[tbp]
  \centering
  \scriptsize
  \caption{Random split of the four public small datasets for training, validation and test.}
    \begin{tabular}{ccccccccc}
    \toprule
    \multirow{2}{*}{Mode} & \multicolumn{2}{c}{Butterfly Mimics} & \multicolumn{2}{c}{Asian vs African Elephants} & \multicolumn{2}{c}{Breast Ultrasound} & \multicolumn{2}{c}{COVID-CT}\\
    \cmidrule(r){2-3} \cmidrule(r){4-5} \cmidrule(r){6-7} \cmidrule(r){8-9}
    &Monarch Butterfly &Viceroy Butterfly &Asian Elephant &African Elephant &Benign &Malignant &Non-COVID-19 &COVID-19   \\
    \midrule
    Train      &90 &67 &294 &296 &440 &158 &234 &191\\
    \midrule
    Validation &21 &21 &100 &100 &53  &53  &58  &60 \\
    \midrule
    Test       &21 &21 &100 &100 &53  &53  &105 &98 \\
    \bottomrule
    \end{tabular}
\label{tab_dataset}
\end{table*}

\section{Implementation Details}
\subsection{Network Architecture}
The generators adopt \cite{resnet-generator} which contains three convolutions for downsampling, nine residual blocks, and two transposed convolutions with the stride size of $\frac{1}{2}$ for upsampling. We adopt instance normalization~\cite{instancenormalization} in the generators but no normalization in the first convolution of discriminator. We add one channel for the first convolutional layer, because the input distance needs to be spatially replicated match the size of the input image and concatenated with the input image. The discriminators use PatchGANs~\cite{pix2pix} to determine if the $70 \times 70$ overlapping image patches is a real one or a generated one.

We randomly select various classifiers with different architectures including AlexNet \cite{DBLP:conf/nips/KrizhevskySH12}, VGG16~\cite{vgg}, GoogleLeNet~\cite{googlenet}, ResNet34~\cite{resnet}, DenseNet \cite{densenet}, MnasNet 1\_0~\cite{mnasnet}, EfficientNet V1 (B5)~\cite{efficientnet} and ConvNeXt (Tiny) \cite{convnext}.
\begin{figure*}[tbp]
	\centering
	\includegraphics[width=\textwidth]{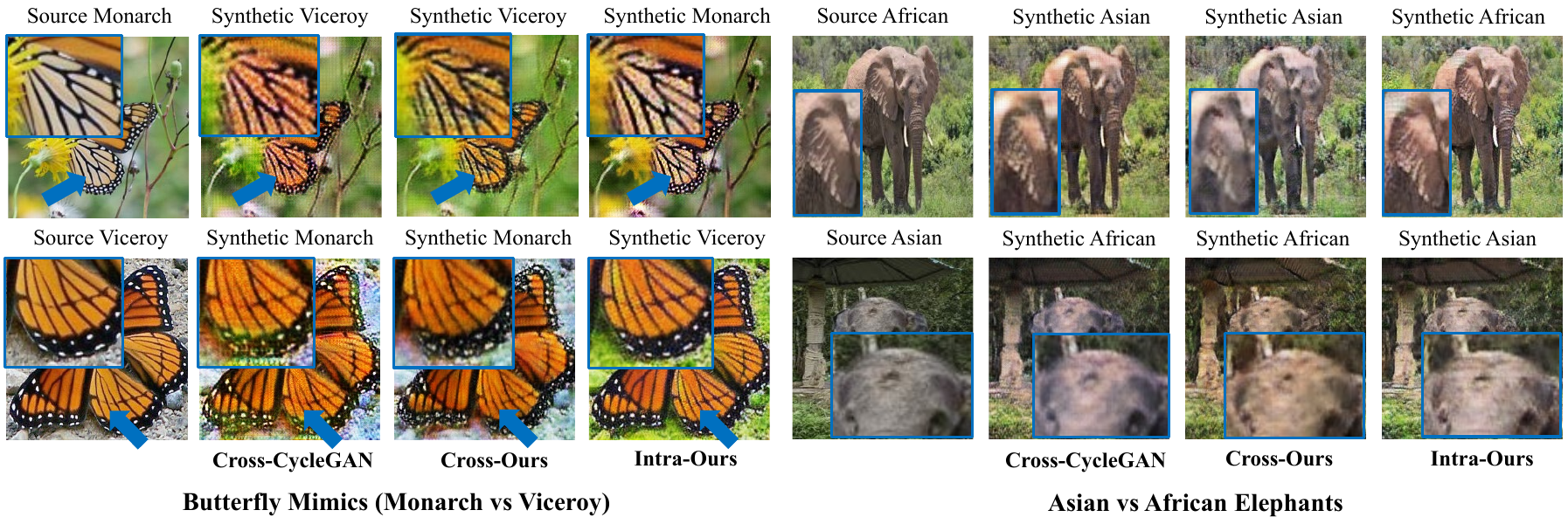}
	\caption{Qualitative results over Butterfly Mimics~\cite{butterfly} and Asian vs African Elephants~\cite{GoumiriBP23}. The generated images by the DisGAN are clearly more realistic than that by the CycleGAN~\cite{cyclegan}. The distance parameters for generating these images are provided by the auxiliary classifier ConvNeXt~\cite{convnext}.}
	\label{fig_NaturalResult}
\end{figure*}
\begin{table*}[tbp]
\centering
\caption{Comparisons with the GAN-based methods over Butterfly Mimics and Asian vs African Elephants. The proposed augmentation method (TA+DisGAN) mostly performs the best. We report ACC ($\uparrow$) averaged over three runs.}
\resizebox{\linewidth}{!}{
    \begin{tabular}{cccccccccc}
    \toprule
    Datasets & Methods & AlexNet & VGG16 & GoogleNet & ResNet34 & DenseNet121 & MnasNet1\_0 & EfficientNet  V1 & ConvNeXt \\
    \midrule
    \multicolumn{1}{c}{\multirow{6}[12]{*}{\makecell{Butterfly Mimics \\ (Monarch vs Viceroy)}}} & Original & 0.817±.009 & 0.899±.014 & 0.889±.055 & 0.802±.050 & 0.857 ±.000 & 0.921±.014 & 0.928±.024 & 0.928±.024 \\
\cmidrule{2-10}          & \makecell{Traditional \\ Augmentation (TA)} & 0.857±.024 & 0.929±.000 & 0.881±.000 & 0.841±.036 & 0.897±.014 & 0.897±.027 & \textbf{0.937±.014} & 0.929±.000 \\
\cmidrule{2-10}          & TA+ACGAN & 0.857±.041 & 0.921±.014 & 0.881±.000 & 0.770±.027 & 0.897±.013 & 0.921±.013 & 0.913±.027 & 0.929±.000 \\
\cmidrule{2-10}          & TA+VACGAN & 0.802±.050 & \textbf{0.936±.014} & 0.857±.041 & 0.722±.060 & 0.857±.041 & 0.897±.014 & 0.873±.077 & 0.913±.014 \\
\cmidrule{2-10}          & TA+CycleGAN & 0.802±.060 & 0.913±.014 & \textbf{0.897±.014} & 0.794±.027 & 0.873±.050 & 0.881±.024 & 0.905±.041 & 0.913±.014 \\
\cmidrule{2-10}          & TA+DisGAN & \textbf{0.881±.024} & 0.929±.024 & 0.841±.027 & \textbf{0.873±.028} & \textbf{0.905±.024} & \textbf{0.926±.014} & 0.929±.000 & \textbf{0.937±.014} \\
    \midrule
    \multicolumn{1}{c}{\multirow{6}[12]{*}{\makecell{Asian vs African\\Elephants}}} & Original & 0.750±.017 & 0.832±.059 & 0.797±.051 & 0.727±.018 & 0.748±.023 & \textbf{0.863±.006} & 0.885±.013 & 0.900±.017 \\
\cmidrule{2-10}          & \makecell{Traditional \\ Augmentation (TA)} & 0.763±.033 & 0.848±.024 & 0.822±.018 & 0.735±.013 & 0.787±.043 & 0.858±.025 & 0.853±.023 & 0.908±.016 \\
\cmidrule{2-10}          & TA+ACGAN & 0.788±.072 & 0.742±.059 & 0.838±.038 & 0.660±.022 & 0.792±.055 & 0.838±.038 & 0.868±.019 & 0.883±.003 \\
\cmidrule{2-10}          & TA+VACGAN & 0.755±.051 & 0.750±.053 & 0.640±.009 & 0.657±.020 & 0.655±.013 & 0.825±.033 & 0.807±.046 & 0.870±.017 \\
\cmidrule{2-10}          & TA+CycleGAN & 0.822±.031 & 0.828±.034 & 0.852±.021 & 0.695±.058 & 0.735±.022 & \textbf{0.863±.006} & 0.767±.012 & 0.918±.006 \\
\cmidrule{2-10}          & TA+DisGAN & \textbf{0.830±.019} & \textbf{0.867±.009} & \textbf{0.865±.008} & \textbf{0.748±.032} & \textbf{0.832±.005} & 0.848±.014 & \textbf{0.886±.032} & \textbf{0.925±.006} \\
    \bottomrule
    \end{tabular}%
}
\label{tab_natural_acc}
\end{table*}
\begin{table*}[tbp]
\centering
\caption{Comparisons with the state of the arts over Butterfly Mimics and Asian vs African Elephants. The proposed augmentation method (TA+DisGAN) mostly performs the best. We report AUC ($\uparrow$) averaged over three runs.}
\resizebox{\linewidth}{!}{
    \begin{tabular}{cccccccccc}
    \toprule
    Datasets & Methods & AlexNet & VGG16 & GoogleNet & ResNet34 & DenseNet121 & MnasNet1\_0 & EfficientNet V1 & ConvNeXt \\
    \midrule
    \multicolumn{1}{c}{\multirow{6}[12]{*}{\makecell{Butterfly Mimics \\ (Monarch vs Viceroy)}}} & Original & 0.921±.023 & 0.955±.007 & 0.945±.011 & 0.923±.020 & 0.927±.034 & \textbf{0.940±.005} & 0.958±.015 & 0.943±.012 \\
\cmidrule{2-10}          & \makecell{Traditional \\ Augmentation (TA)}   & 0.910±.018 & 0.940±.014 & 0.945±.003 & 0.931±.022 & 0.954±.015 & 0.938±.026 & 0.960±.006 & 0.960±.008 \\
\cmidrule{2-10}          & TA+ACGAN    & 0.924±.018 & 0.959±.015 & 0.941±.017 & 0.901±.043 & 0.966±.009 & 0.936±.004 & 0.955±.016 & 0.953±.012 \\
\cmidrule{2-10}          & TA+VACGAN    & 0.876±.051 & 0.950±.008 & 0.933±.001 & 0.864±.048 & 0.936±.025 & 0.936±.006 & 0.930±.018 & 0.959±.004 \\
\cmidrule{2-10}          & TA+CycleGAN    & 0.924±.004 & 0.962±.014 & 0.948±.006 & 0.915±.001 & 0.919±.024 & 0.938±.009 & 0.944±.035 & 0.936±.005 \\
\cmidrule{2-10}          & TA+DisGAN    & \textbf{0.932±.014} & \textbf{0.963±.007} & \textbf{0.952±.014} & \textbf{0.938±.015} & \textbf{0.969±.009} & 0.938±.010 & \textbf{0.961±.006} & \textbf{0.966±.006} \\
    \midrule
    \multicolumn{1}{c}{\multirow{6}[12]{*}{\makecell{Asian vs African\\Elephants}}} & Original   & 0.845±.028 & 0.897±.050 & 0.873±.032 & 0.806±.040 & 0.850±.036 & 0.933±.007 & 0.921±.005 & 0.948±.031 \\
\cmidrule{2-10}          & \makecell{Traditional \\ Augmentation (TA)}  & 0.861±.010 & 0.915±.016 & 0.900±.010 & 0.811±.019 & 0.905±.026 & 0.937±.005 & 0.906±.002 & 0.948±.020 \\
\cmidrule{2-10}          & TA+ACGAN   & 0.860±.061 & 0.828±.078 & 0.922±.027 & 0.706±.052 & 0.862±.046 & 0.933±.009 & \textbf{0.934±.009} & 0.938±.000 \\
\cmidrule{2-10}          & TA+VACGAN    & 0.855±.040 & 0.842±.015 & 0.712±.015 & 0.733±.015 & 0.751±.041 & 0.931±.010 & 0.886±.013 & 0.928±.009 \\
\cmidrule{2-10}          & TA+CycleGAN   & 0.896±.010 & 0.896±.014 & \textbf{0.924±.013} & 0.743±.060 & 0.790±.022 & 0.940±.003 & 0.843±.013 & 0.966±.006 \\
\cmidrule{2-10}          & TA+DisGAN   & \textbf{0.898±.005} & \textbf{0.925±.008} & \textbf{0.924±.001} & \textbf{0.838±.028} & \textbf{0.917±.018} & \textbf{0.944±.007} & 0.927±.015 & \textbf{0.969±.004} \\
    \bottomrule
    \end{tabular}%
}
\label{tab_natural_auc}
\end{table*}

\subsection{Training Settings}
All networks are trained using Adam~\cite{adam} with $\beta_1=0.5$ and $\beta_2=0.999$. The initial learning rate is set to $1\times e^{-4}$ over the first 25 epochs and linearly decays to 0 over the next 25 epochs. All images are normalized between -1 and 1, and resized to $224\times224$. For all classifiers optimized by hinge loss, the number of output node of last linear layer is set to 1. We adopt transfer learning for all classifiers using ImageNet Dataset~\cite{imagenet}. We use traditional augmentation to pre-augment the training samples for all experiments.

We set the same classification weights for the real samples and that of the generated samples. For fair comparison, we follow the CycleGAN \cite{cyclegan} for $\lambda_{\rm interCYC} = 10.0$, $\lambda_{\rm intraCYC} = 10.0$. Considering the vertical distances and horizontal distances having different large-scales, the hyper-parameter $\lambda_{\rm verDIS}$ is set from 0.01 to 0.1 and the hyper-parameter $\lambda_{\rm horDis}$ is set from 0.001 to 0.01. The details of hyper-parameters are described in Table~\ref{tab_hyperparameters} for better reproducibility. Note that the $\lambda_{\rm horDIS}$ is less than or equal to the $\lambda_{\rm verDIS}$, and several $\lambda_{\rm horDIS}$, $\lambda_{\rm verDIS}$ of classifiers are less or equal to that of other classifiers. These settings have the following two reasons respectively: (1) The error of reconstructing the horizontal distances is larger than that of reconstructing the vertical distances. (2) Several specific auxiliary classifiers constructs overfitting hyperplanes resulting in the larger scales of the distances.

\section{Experiments and Results}
\subsection{Datasets Description and Evaluation Metrics}
We conduct experiments over multiple public datasets: Butterfly Mimics~\cite{butterfly}, Asian vs African Elephants~\cite{GoumiriBP23, Data-centric}, Breast Ultrasound Images Dataset (BUSI)~\cite{BUSI}, Breast Ultrasound images Collected from UDIAT Diagnostic Center (UDIAT)~\cite{UDIAT} and CT Scan Dataset about COVID-19 (COVID-CT)~\cite{zhao2020COVID-CT-Dataset}, and perform evaluations with two widely adopted metrics in image classification: accuracy (ACC) and the area under the receiver operating characteristic curve (AUC). We mix the BUSI dataset with UDIAT dataset to increase the challenge of breast ultrasound image classification.

For all the experiments, four public small datasets are used: Butterfly Mimics, Asian vs African Elephants, mixed Breast Ultrasound and COVID-CT. Butterfly Mimics contains 132 monarch butterfly images and 109 Viceroy butterfly images. Asian vs African Elephants contains 494 asian elephant images and 496 african elephant images. mixed Breast Ultrasound contains 546 benign breast lesion images and 264 malignant breast lesion images. COVID-CT contains 307 non-COVID-19 images and 349 COVID-19 CT images. For each of the dataset, we randomly split them for training, validation and test. The details are listed in Table~\ref{tab_dataset}.

\subsection{Experiments on Butterfly Mimics and Elephants}
Figure~\ref{fig_NaturalResult} qualitatively demonstrates that the DisGAN outperforms the inter-domain translation technique (\emph{i.e.,} CycleGAN) in data-limited generation, especially in terms of the generated textures and shapes. The DisGAN is prone to generate the black postmedian stripe across hindwing for viceroy butterflies and eliminate it in the opposite generation direction. The DisGAN generates asian elephants with light gray and african elephants with grayish brown. Moreover, it clears two bumps in the top of the source asian elephant's head.

Table~\ref{tab_natural_acc} and Table \ref{tab_natural_auc} compare the proposed augmentation method with the GAN-based augmentation methods in data-limited natural image classification over the Butterfly Mimics and Asian vs African Elephants. We can see that TA+DisGAN method mostly performs the best in AUC score, demonstrating its effectiveness of improving the classification generalization. Specially, transfer learning based on ImageNet Database~\cite{imagenet} is used to train and fix the binary classifier into an auxiliary classifier in VACGAN and DisGAN. Each row (TA) reflects the classification performance of fixed classifiers.
\begin{figure*}[tbp]
	\centering
	\includegraphics[width=\textwidth]{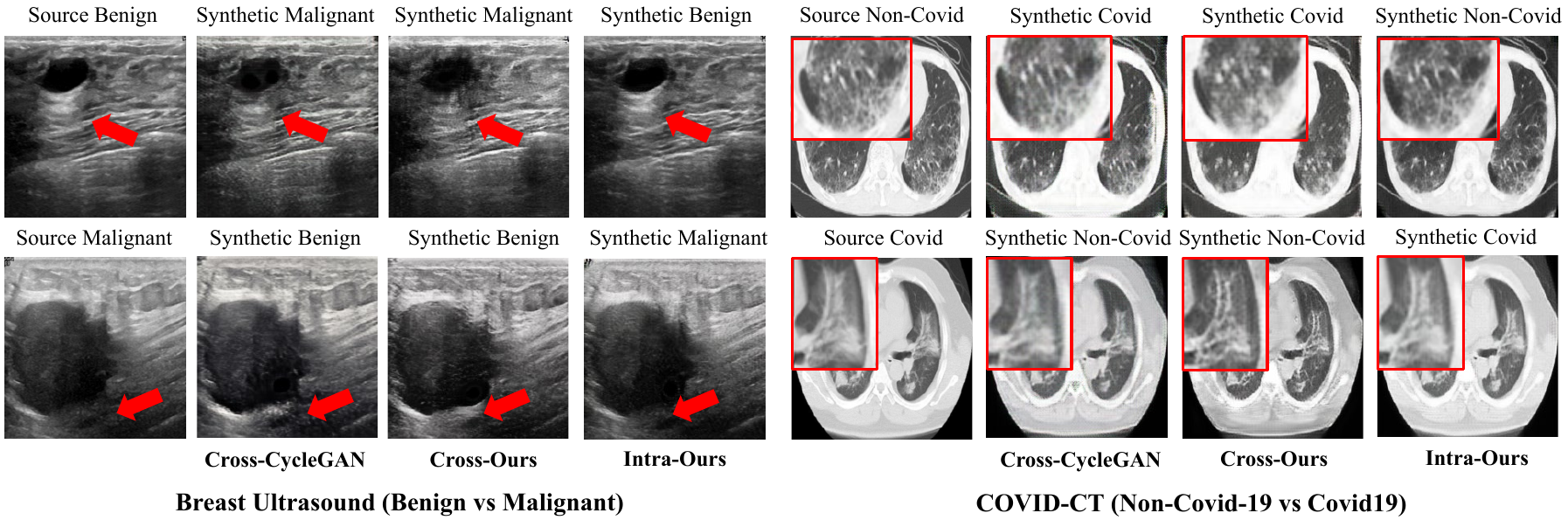}
	\caption{Qualitative results over mixed Breast Ultrasound~\cite{BUSI, UDIAT} and COVID-CT~\cite{zhao2020COVID-CT-Dataset}. The generated images by the DisGAN is clearly more realistic than that by the CycleGAN~\cite{cyclegan}. The distance parameters for generating these images are provided by the auxiliary classifier ConvNeXt~\cite{convnext}. }
	\label{fig_MedicalResult}
\end{figure*}
\begin{figure}[tbp]
	\centering
	\includegraphics[width=\columnwidth]{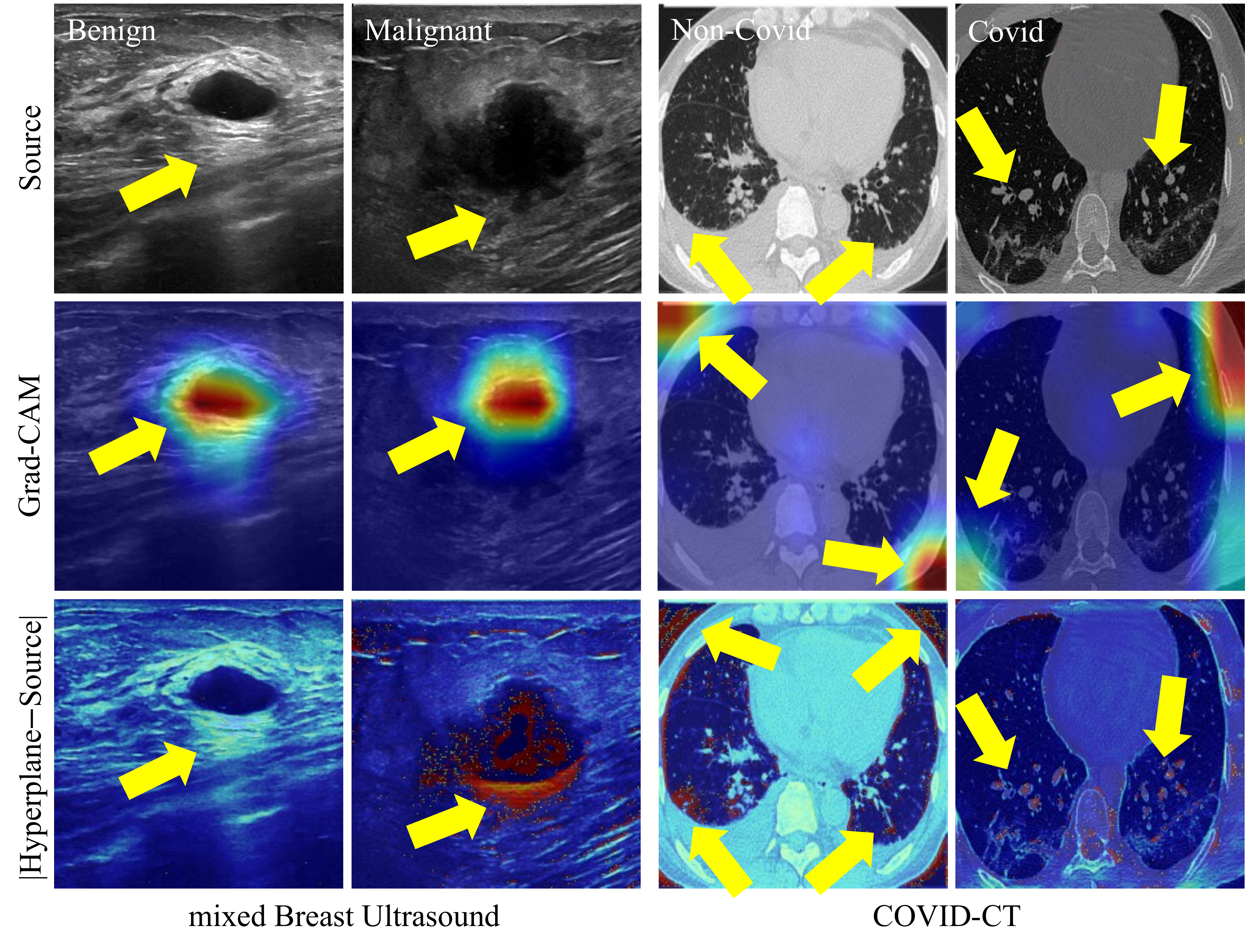}
	\caption{Comparing DisGAN's class-difference maps (CDMs) with Grad-CAM~\cite{gradcam} (TA + CrossEntroy) over the mixed Breast Ultrasound and the COVID-CT datasets. The difference between the source images and their projections on a hyperplane can highlight class-specific region, which cannot be deduced from the Grad-CAM~\cite{gradcam}.}
	\label{fig_Interpretability}
\end{figure}
\begin{table*}[tbp]
\centering
\caption{Comparisons with the state of the arts over mixed Breast Ultrasound and COVID-CT. The proposed augmentation method (TA+DisGAN) mostly outperforms the GAN-based models on the limited medical image datasets. We report ACC ($\uparrow$) averaged over three runs.}
\resizebox{\linewidth}{!}{
    \begin{tabular}{cccccccccc}
    \toprule
    \multicolumn{1}{c}{Datasets} & Methods & AlexNet & VGG16 & GoogleNet & ResNet34 & DenseNet121 & MnasNet1\_0 & EfficientNet  V1 & ConvNeXt \\
    \midrule
    \multicolumn{1}{c}{\multirow{6}[12]{*}{\makecell{Breast Ultrasound \\ (Benign vs Malignant)}}} & Original   & 0.833±.005 & 0.821±.000 & 0.830±.028 & \textbf{0.830±.009} & 0.789±.024 & 0.852±.005 & 0.843±.014 & 0.852±.029 \\
\cmidrule{2-10}          & \makecell{Traditional \\ Augmentation (TA)}    & 0.805±.027 & 0.763±.025 & 0.811±.052 & 0.815±.038 & 0.802±.028 & 0.824±.045 & 0.865±.011 & 0.862±.005 \\
\cmidrule{2-10}          & TA+ACGAN    & 0.821±.000 & 0.755±.050 & 0.774±.059 & 0.783±.009 & 0.792±.016 & 0.840±.019 & 0.774±.087 & 0.874±.011 \\
\cmidrule{2-10}          & TA+VACGAN    & 0.805±.005 & 0.774±.049 & 0.758±.044 & 0.767±.093 & 0.641±.086 & 0.830±.028 & 0.704±.080 & 0.868±.016 \\
\cmidrule{2-10}          & TA+CycleGAN    & 0.805±.005 & 0.821±.025 & 0.811±.028 & 0.805±.014 & 0.774±.041 & 0.833±.014 & 0.849±.016 & 0.874±.005 \\
\cmidrule{2-10}          & TA+DisGAN    & \textbf{0.859±.019} & \textbf{0.830±.011} & \textbf{0.852±.014} & 0.821±.028 & \textbf{0.837±.014} & \textbf{0.886±.014} & \textbf{0.878±.014} & \textbf{0.887±.009} \\
    \midrule
    \multicolumn{1}{c}{\multirow{6}[12]{*}{\makecell{COVID-CT \\ (Non-Covid-19 \\ vs \\ Covid-19)}}} & Original    & 0.734±.020 & 0.765±.014 & 0.691±.058 & 0.713±.021 & 0.727±.012 & 0.777±.027 & \textbf{0.783±.039} & 0.777±.030 \\
\cmidrule{2-10}          & \makecell{Traditional \\ Augmentation (TA)}    & 0.724±.037 & 0.736±.032 & 0.755±.040 & 0.695±.013 & 0.757±.010 & \textbf{0.803±.031} & 0.754±.063 & 0.793±.010 \\
\cmidrule{2-10}          & TA+ACGAN    & 0.657±.059 & 0.773±.023 & 0.667±.047 & 0.714±.015 & 0.718±.057 & 0.731±.006 & 0.701±.078 & 0.803±.000 \\
\cmidrule{2-10}          & TA+VACGAN    & 0.673±.017 & 0.698±.069 & 0.696±.038 & 0.619±.088 & 0.652±.029 & 0.750±.003 & 0.654±.010 & 0.780±.017 \\
\cmidrule{2-10}          & TA+CycleGAN    & 0.709±.027 & 0.757±.033 & 0.765±.006 & 0.732±.032 & 0.742±.037 & 0.771±.006 & 0.745±.019 & 0.775±.022 \\
\cmidrule{2-10}          & TA+DisGAN    & \textbf{0.765±.010} & \textbf{0.789±.008} & \textbf{0.780±.019} & \textbf{0.770±.019} & \textbf{0.789±.009} & 0.793±.013 & \textbf{0.783±.008} & \textbf{0.816±.012} \\
    \bottomrule
    \end{tabular}%
}
\label{tab_medical_acc}
\end{table*}
\begin{table*}[tbp]
\centering
\caption{Comparisons with the state of the arts over mixed Breast Ultrasound and COVID-CT. The proposed augmentation method (TA+DisGAN) mostly outperforms the GAN-based models on the limited medical image datasets. We report AUC ($\uparrow$) averaged over three runs.}
\resizebox{\linewidth}{!}{
    \begin{tabular}{cccccccccc}
    \toprule
    \multicolumn{1}{c}{Datasets} & Methods & AlexNet & VGG16 & GoogleNet & ResNet34 & DenseNet121 & MnasNet1\_0 & EfficientNet  V1 & ConvNeXt \\
    \midrule
    \multicolumn{1}{c}{\multirow{6}[12]{*}{\makecell{Breast Ultrasound \\ (Benign vs Malignant)}}} & Original    & 0.931±.011 & 0.910±.014 & 0.910±.013 & 0.910±.013 & 0.892±.024 & 0.925±.007 & 0.933±.007 & 0.906±.020 \\
\cmidrule{2-10}          & \makecell{Traditional \\ Augmentation (TA)}    & 0.897±.008 & 0.907±.002 & 0.887±.048 & 0.900±.015 & 0.891±.028 & 0.920±.011 & 0.915±.012 & 0.935±.005 \\
\cmidrule{2-10}          & TA+ACGAN    & 0.917±.025 & 0.867±.020 & 0.875±.027 & 0.867±.036 & 0.896±.003 & 0.913±.007 & 0.853±.089 & 0.949±.003 \\
\cmidrule{2-10}          & TA+VACGAN   & 0.920±.003 & 0.885±.017 & 0.854±.014 & 0.847±.046 & 0.813±.063 & 0.932±.014 & 0.840±.022 & 0.939±.010 \\
\cmidrule{2-10}          & TA+CycleGAN    & 0.894±.015 & 0.922±.018 & 0.914±.009 & 0.889±.019 & 0.867±.013 & 0.911±.017 & 0.919±.011 & 0.946±.006 \\
\cmidrule{2-10}          & TA+DisGAN    & \textbf{0.931±.001} & \textbf{0.929±.010} & \textbf{0.919±.005} & \textbf{0.920±.003} & \textbf{0.922±.016} & \textbf{0.934±.007} & \textbf{0.940±.010} & \textbf{0.953±.009} \\
    \midrule
    \multicolumn{1}{c}{\multirow{6}[12]{*}{\makecell{COVID-CT\\(Non-Covid-19 \\ vs \\ Covid-19)}}} & Original    & 0.793±.013 & 0.842±.015 & 0.764±.077 & 0.770±.003 & 0.818±.018 & 0.830±.020 & 0.831±.042 & 0.868±.020 \\
\cmidrule{2-10}          & \makecell{Traditional \\ Augmentation (TA)}    & 0.789±.017 & 0.833±.024 & 0.818±.037 & 0.750±.016 & 0.816±.012 & 0.835±.008 & 0.837±.036 & 0.876±.004 \\
\cmidrule{2-10}          & TA+ACGAN    & 0.724±.051 & 0.849±.011 & 0.780±.032 & 0.764±.019 & 0.776±.063 & 0.787±.001 & 0.772±.083 & 0.877±.001 \\
\cmidrule{2-10}          & TA+VACGAN    & 0.775±.025 & 0.813±.074 & 0.780±.034 & 0.708±.053 & 0.741±.044 & 0.818±.006 & 0.738±.036 & 0.849±.003 \\
\cmidrule{2-10}          & TA+CycleGAN    & 0.784±.010 & 0.830±.034 & \textbf{0.856±.011} & 0.791±.025 & 0.805±.032 & 0.835±.002 & 0.819±.016 & 0.838±.012 \\
\cmidrule{2-10}          & TA+DisGAN    & \textbf{0.813±.010} & \textbf{0.852±.006} & 0.842±.019 & \textbf{0.838±.013} & \textbf{0.845±.010} & \textbf{0.837±.008} & \textbf{0.844±.007} & \textbf{0.885±.013} \\
    \bottomrule
    \end{tabular}%
}
\label{tab_medical_auc}
\end{table*}

\subsection{Experiments on Breast Ultrasound and COVID-CT}
Figure \ref{fig_MedicalResult} demonstrates that the DisGAN can generate more realistic images compared with the CycleGAN. The DisGAN generates dark shadowing below the lesion when transform the source benign images, and generate bright region in the inverse translation. Moreover, we can observe the smoothness changes of the breast lesion's boundary. In bi-directional generation of CT images, we can also see the generation and elimination of ground glass shadow regions.

Table \ref{tab_medical_acc} and Table \ref{tab_medical_auc} quantitatively compares the proposed augmentation method with the GAN-based data augmentation method over datasets Mixed Breast Ultrasound and COVID-CT. We can see that the TA+DisGAN method consistently outperforms traditional augmentation and mostly outperforms the GAN-based data augmentation in AUC score, especially using the limited training samples.

\begin{table}
    \centering
    \caption{Ablation study of the DisGAN: VerDisGAN, HorDisGAN and DisGAN. The DisGAN performs the best as VerDisGAN and HorDisGAN are complementary to each other. We report the the ACC ($\uparrow$) and AUC ($\uparrow$) averaged over the eight classifiers.}
    \resizebox{\linewidth}{!}{
        \begin{tabular}{cccccc}
        \toprule
        DataSets & Metrics & \multicolumn{1}{c}{\makecell{Traditional\\Augmentation\\(TA)}} & \multicolumn{1}{c}{\makecell{TA+\\VerDisGAN}} & \multicolumn{1}{c}{\makecell{TA+\\HorDisGAN}} & \multicolumn{1}{c}{\makecell{TA+\\VerDisGAN+ \\HorDisGAN}} \\
        \midrule
        Butterfly Mimics & ACC   & 0.896 & 0.901 & 0.893 & \textbf{0.902} \\
        \makecell{Asian vs African \\ Elephants} & ACC   & 0.821 & 0.842 & 0.827 & \textbf{0.850} \\
        Breast Ultrasound & ACC   & 0.818 & 0.840 & 0.823 & \textbf{0.856} \\
        COVID-CT & ACC   & 0.752 & 0.770 & 0.759 & \textbf{0.786} \\
        \midrule
        Butterfly Mimics & AUC   & 0.942 & 0.946 & 0.940 & \textbf{0.952} \\
        \makecell{Asian vs African \\ Elephants} & AUC   & 0.898 & 0.896 & 0.908 & \textbf{0.918} \\
        Breast Ultrasound & AUC   & 0.907 & 0.920 & 0.911 & \textbf{0.931} \\
        COVID-CT & AUC   & 0.819 & 0.831 & 0.826 & \textbf{0.845} \\
        \bottomrule
        \end{tabular}%
    }
    \label{tab_ablation}
\end{table}
\begin{figure*}[tbp]
	\centering
	\includegraphics[width=\textwidth]{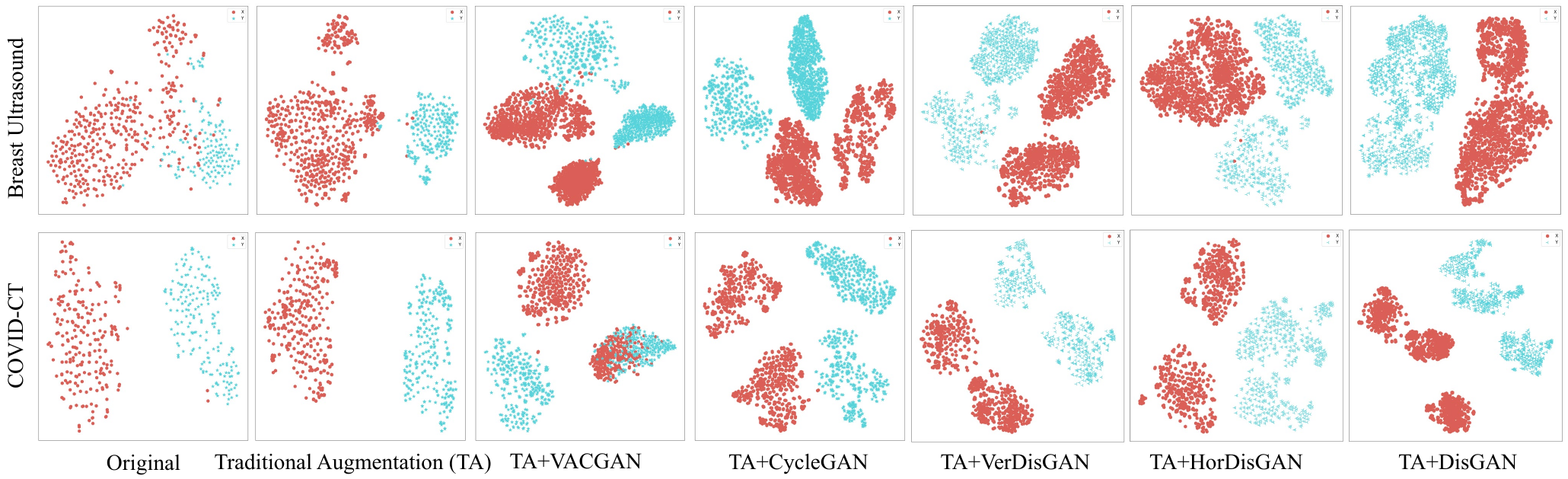}
	\caption{Training samples' distribution of ConvNeXt~\cite{convnext} visualized by t-SNE. From left to right column: original samples, traditional augmentation (TA), TA+VACGAN, TA+CycleGAN, TA+VerDisGAN, TA+HorDisGAN, TA+DisGAN. We observer that TA+DisGAN can construct a more precise hyperplane by generating samples alongside the its margins.}
	\label{fig_ConvNeXtTSNE}
\end{figure*}

\subsection{Interpretability of Classification}
The proposed DisGAN provides the class-difference maps (CDMs) for explaining the binary classifiers. Figure~\ref{fig_Interpretability} shows that the GradCAM \cite{gradcam} focuses on the regions with large gradient changes in each class image, whereas the proposed class-difference maps (CDMs) can focus regions with large changes between each image and its projection on hyperplane. We define the CDMs for images in domain $X$ and domain $Y$ using $|x - G_{X2Y}(x, 0))|$ and $|y - G_{Y2X}(y, 0))|$ respectively.

\subsection{Ablation Study}
The DisGAN consists of two major components: the VerDisGAN which is conditioned on the vertical distances and the HorDisGAN which is conditioned on the vertical distances. We evaluate the contributions of these two components to the overall classification respectively. As illustrated in Table~\ref{tab_ablation}, DisGAN gets the highest ACC and AUC scores using the average performance of the eight classification architectures. Moreover, VerDisGAN makes more contributions to improving the classification generalization than HorDisGAN.

\subsection{Visualization of Training Sample Distribution}
\label{convnext_distribution}
The generated samples of DisGAN participate in reshaping the hyperplane in the hyperplane space. Figure~\ref{fig_ConvNeXtTSNE} shows the training samples' distribution of GAN-based augmentation methods using t-Distributed Stochastic Neighbor Embedding (t-SNE) \cite{tsne} with ConvNeXt. VACGAN converts noise-vector to images with condition of binary class labels. CycleGAN translates source image to target image with only image space constraint. These two methods lead to the label uncertainty of generated samples. Compared with the previous GAN-based method, DisGAN translates source images to target images with both the image space constraint and the hyperplane distance constraint.
\begin{figure*}[tbp]
	\centering
	\includegraphics[width=\textwidth]{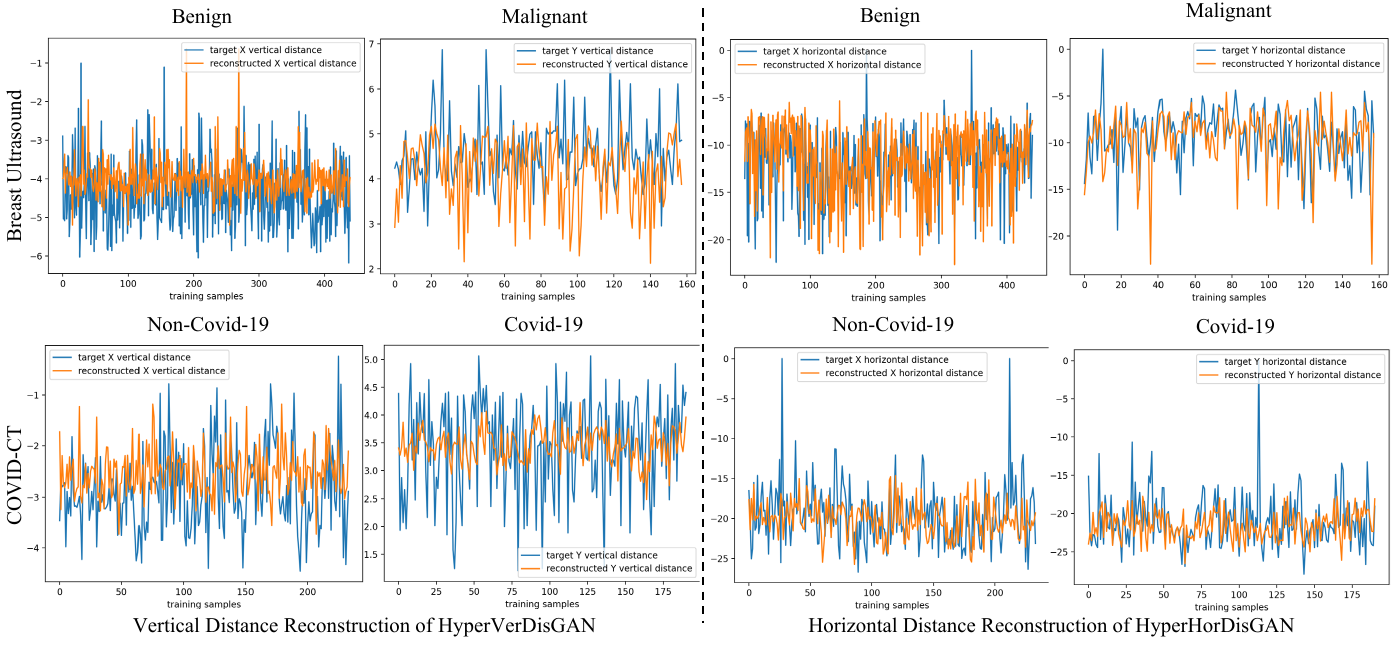}
	\caption{The curves of vertical and horizontal distances (Blue) and the auxiliary classifier's reconstructed distances (Orange). We show this example using classifier ConvNeXt.}
	\label{fig_DistanceMedical}
\end{figure*}

\section{Discussion}
\subsection{Innovation of Parameterized Network}
To the best of our knowledge, we are the first to control the variation degrees of generate samples with respect to the classification decision boundary, which can reshape the boundary for the binary classification task. The variation degrees are represented by the vertical distances between the target samples and the auxiliary classifer's optimal hyperplane, and the horizontal distances between the intra-domain source and target samples.

The SGAN and ACGAN's auxiliary classifiers share the same architectures with the discriminators, whereas the DisGAN's auxiliary classifier is external and can easily apply to any latest architecture. Moreover, SGAN's auxiliary classifier takes the generated samples as an additional class. ACGAN and VACGAN's auxiliary classifiers reconstruct the classification labels from the generated samples. Unlike these methods, DisGAN's auxiliary classifier reconstructs the input distances from the generated samples.
\begin{figure*}[tbp]
	\centering
	\includegraphics[width=\textwidth]{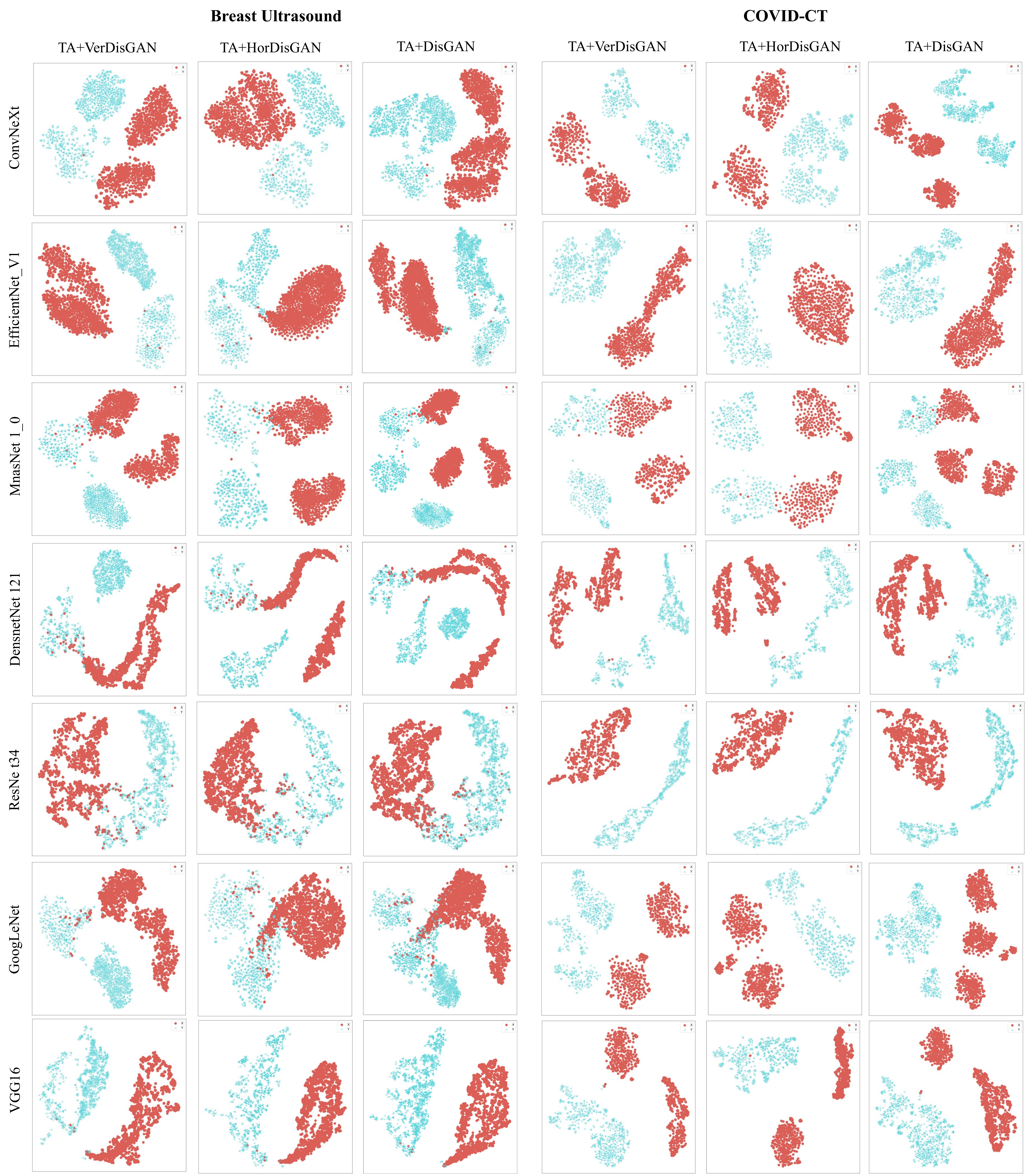}
	\caption{Training samples' distribution of the various classification models~\cite{convnext} visualized by t-SNE \cite{tsne}. From up to low row with different classifiers: ConvNeXt, EfficientNet V1, MnasNet 1\_0, DenseNet 121, ResNet34, GoogLeNet and VGG16.}
	\label{fig_AllTSNE}
\end{figure*}

\subsection{Effectiveness of Parameterized Network}
The success of the DisGAN is the controllable variation degrees represented by the vertical and horizontal distances. The ability of the distance guided generators are shown in Figure~\ref{fig_DistanceMedical} on two medical datasets: the reconstructed vertical distance curve trend is generally consistent with the target's vertical distance curve (left); and the reconstructed horizontal distance curve trend is consistent with the target's horizontal distance curve (right). Therefore, controlling the variation degrees of generated samples can be achieved by the distance guided generation, whereas the previous GAN-based methods are difficult to achieve this goal.

\subsection{Training Data Distribution for State-of-the-arts Classifiers}
We have shown the training data distribution for ConvNeXt using two limited medical datasets in Section \ref{convnext_distribution}. Now we show training data distribution for other classification models. Figure \ref{fig_AllTSNE} compares the distribution of traditional augmentation (TA) + DisGAN on the two limited medical datasets by t-Distributed Stochastic Neighbor Embedding (t-SNE)~\cite{tsne}. The DisGAN with different auxiliary classifiers tend to fill the hyperplane spaces of these classification models, and generate samples along the margins of the hyperplanes.

\subsection{Extending to Multi-class Classification}
The current method has potential to be extended to multi-class classification. We can add distance constraints into starGAN \cite{stargan}, which use one generator to take the soure image and the target domain label as input. A $N$-class classifier by using multi-hinge loss can be trained. Each output node of the last fully connection layer representing a binary classifier's output. The $i^{th}$ binary classifier aims to classify between $i^{th}$ domain and not $i^{th}$ domain.  We can measure the two types distances for the source sample in $i^{th}$ domain. One is a vertical distance from a target domain sample to $i^{th}$ optimal hyperplane, and another is the horizontal distance from a target intra-domain sample to the source sample. In this direction, we can generate the intra-domain samples which are conditioned on the horizontal distances and target domain label $i \in N$; and generate the inter-domain samples which are conditioned on the vertical distances and target domain label $j (j \neq i) \in N$.

\section{Conclusion}
The primary aim of this paper is to develop a general data augmentation method that can be applied to different classification architectures. We have showcased the effectiveness of the DisGAN for generating diverse samples to improve the various classification models in both natural and medical datasets. We observe that controlling the variation degrees of generated samples has significantly more impact than blindly increasing the number of the training samples. In addition, difference between the source samples and their mappings on decision boundary can contribute explanation.

The current work has limitations that need to be studied in the future: (1) The DisGAN primarily focuses on the binary classifications, and exploiting hyperplane between multi-classes remains to be investigated. (2) Controlling the variation degrees of generated samples utilizes the fixed auxiliary classifier to construct the hyperplane. We hypothesize that this can be an interactive process: the binary classifier which uses generated samples can be used again as the fixed auxiliary classifier. This iterative process is expected to be the part where the artificial intelligence improves itself. (3) A comprehensive evaluation of the generated samples in terms of usefulness beyond augmenting training samples.

\bibliographystyle{IEEEtran}
\bibliography{tnnls}

\vfill

\end{document}